\documentclass{article}

\usepackage[nonatbib,preprint]{neurips_2023}

\usepackage[utf8]{inputenc} 
\usepackage[T1]{fontenc}    
\usepackage{hyperref}       
\usepackage{url}            
\usepackage{booktabs}       
\usepackage{amsfonts}       
\usepackage{nicefrac}       
\usepackage{microtype}      
\usepackage{graphicx}
\usepackage{amsmath}
\usepackage{float}
\usepackage[english]{babel}

\usepackage{booktabs}
\usepackage[table]{xcolor}
\usepackage{dsfont}
\usepackage{caption}
\usepackage{subcaption}
\usepackage{paralist}

\renewcommand\vec{\mathbf}

\newcommand{\R}{\mathbb{R}}
\DeclareMathOperator{\argmax}{arg\,max} 

\def\stoptable#1{%
	\par\vspace{\abovecaptionskip}%
	{\footnotesize #1}%
}

\title{Fantastic DNN Classifiers and How to Identify them without Data}

\author{%
  Nathaniel Dean\\
  Department of Computer Science\\
  University of Miami\\
  Coral Gables, FL \\
  \texttt{nxd551@miami.edu} \\
  \And
  Dilip Sarkar \\
  Department of Computer Science \\
  University of Miami \\
  Coral Gables, FL \\
  \texttt{sarkar@cs.miami.edu} \\
}

\begin{document}

\maketitle

\begin{abstract}
Current algorithms and architecture can create excellent DNN classifier models from example data. In general, larger training datasets result in better model estimations, which improve test performance. Existing methods for predicting generalization performance are based on hold-out test examples. To the best of our knowledge, at present no method exists that can estimate the quality of a trained classifier without test data.
In this paper, we show that the quality of a trained DNN classifier can be assessed without any example data. We consider DNNs to be composed of a feature extractor and a feature classifier; the feature extractor’s output is fed to the classifier. The proposed method iteratively creates class prototypes in the input space for each class by minimizing a cross-entropy loss function at the output of the network.  We use these prototypes and their feature relationships to reveal the quality of the classifier. We have developed two metrics: one using the features of the prototypes and the other using adversarial examples corresponding to each prototype. Empirical evaluations show that accuracy obtained from test examples is directly proportional to quality measures obtained from the proposed metrics. We report our observations for ResNet18 with Tiny ImageNet, CIFAR100, and CIFAR10 datasets. The proposed metrics can be used to compare performances of two or more trained classifiers without testing examples.\footnote{Code available at \url{https://github.com/UMMLearn/FantasticNetworksNoData}.}
\end{abstract}

\section{Introduction}\label{Intro}

As the pace of machine learning development accelerates, more and more users, including private enterprises, are incorporating artificial intelligence systems into their organizations.  With current deep learning methods, the expertise of these systems rely heavily on the scale of both the model and the dataset.  However, end-users of these models may have limited knowledge of the amount or quality of data used to train and test them, particularly in cases where data is third-party proprietary, privacy sensitive, or difficult to analyze due to scale.  If the published accuracies of two models cannot be validated because the test data is not publicly released, is there another metric we can use to compare them?  We propose it would be useful to create an assessment method for deep learning models that is based solely on their parameters and intended task as an alternative to a data-based evaluation when it is infeasible.



\paragraph{Problem Statement}\label{Problem}

We are given a trained neural network for the classification of images. Our objective is to assess the network's accuracy performance \emph{without} having access to a either the training or test set originally used to supervise the network.


\paragraph{Background and Related Work}  
\label{sec:BackgroundAndRelatedWork}
\par
Often Machine Learning (ML) models are trained, validated, and tested with example data. A significant part of the data is used to estimate the parameters of a model, next validation data is used for hyperparameter and model selection, and finally the test data is used to evaluate performance of the selected model \cite{ModelEvaluationModelSelection2020arXiv,EvaluatingMachineLearningModels2019eBook}. 
Using this recipe, Deep Neural Network (DNN) models have shown excellent performance on test data, but further inspection has revealed trouble transferring this accuracy metric to unseen out-of-distribution \cite{nagarajan2021OOD} and adversarial examples \cite{chakraborty2018adversarial}, implying that hold-out set test accuracy is \emph{one} measure of model quality.

The community and the NeurIPS conference have shown interest in establishing new measures that predict generalization gap and offer insight into how DNNs generalize including a NeurIPS 2020 competition \cite{NeurIPS2020CompetetionarXiv}.  The competition asked for new computable measures on given model and training set combinations that correlated well with their generalization gaps, which led to solutions like \cite{natekar2020representation} and later-on \cite{PredictingDeepNeuralNetworkGeneralization2021arXiv}.  Inherently, these solutions create measures from relationships between training data examples and the encoded features of the networks.  In our approach, we derive measures based only on the network parameters, as a key element of our contributions is that data is not required.

\paragraph{Class Prototypes in Other Fields}

An implicit belief in deep learning is that successful architectures are able to identify common semantic features from each class that distinguish them from each other.  These important class features can then be captured into data structures called class prototypes.  



The class prototype idea has been studied in multiple areas such as robustness \cite{mustafaRestrictedHiddenSpace}, explainable AI \cite{liPrototype, chenProtoPNet}, distance-based classification \cite{guerrieroDNCMC, mensinkDBIC, karlinskyDML,rippelMagnet}, few-shot learning \cite{snellPrototypeNetworksFewShot}, and continual learning \cite{maiContrasticReplay}.  

However, these lines of research usually compute prototypes directly from dataset examples for a specific machine learning task.  Our methodology assumes we have no access to data, so we derive prototypes from the model parameters and a loss function at the output of the model.  Because the model parameters reflect the outcome of training the model on real data, our prototypes are indirect representations of the training set.



  


\paragraph{Training Set Reconstruction}

Recent work \cite{buzaglo2023,haim2022} has successfully reconstructed training examples by optimizing a reconstruction loss function that requires only the parameters of the model and no data.  An underlying hypothesis of this work is that neural networks encode specific examples from the training set in their model parameters.  In our work, we utilize this concept in a different way by optimizing a cross-entropy loss at the output of the model and constructing class prototypes for each class, also without any data.


The main contributions of our work are:

\begin{itemize}
\item We have proposed and evaluated two metrics for dataless evaluation of DNNs for classifications; to be more clear, our proposed method requires only the DNN's architecture and parameters. One of the metrics can be used to measure classification accuracy and the other can be used to measure robustness against adversarial attacks.
	\item We have proposed a method that uses a given DNN to create prototype examples for each class-category (by iterative back-propagation of classification error), which are then used to observe activations of neurons at the feature layer for computing values of the metrics. 
	\item We have validated the quality of the proposed metrics by measuring classification performance and robustness of DNNs models trained with CIFAR10, CIFAR100, and Tiny ImageNet datasets; the models had ResNet18 architecture.

\end{itemize}

\section{Dataless Methods to Evaluate Performances of DNNs }
\label{DatalessPerformanceEvalation}

\subsection{Notations} 
\label{Notations}   
We are given a differentiable classifier $f(\cdot\ ; \theta)$ that maps an input vector $\vec{x}\in \R^N$ to a vector $\vec{\hat{y}} \in \R^K$, where each element $f_k(\vec{x} ; \theta)=\hat{y}_k$ of $\vec{\hat{y}}$ represents the probability $\vec{x}$ belongs to class k such that $\sum_{k=1}^{K} \hat{y}_k = 1$.  Correspondingly, for any given $\vec{x}$, which possess a true label $y$, the classifier could generate $K$ possible class assignments and is correct only when $y = \argmax_k(\vec{\hat{y}})$.  To associate an input, variable, or label to a true class $k$, we may superscript it with $k$. 

For our purposes, we parse this standard definition of a classifier into the composition of a \emph{feature extraction} function $g:\R^N \rightarrow \R^D$ and a \emph{feature classification} function $h: \R^D \rightarrow \R^K$.  Therefore, $\vec{\hat{y}} = f(\vec{x};\theta) = h(g(\vec{x}; \theta_{g}) ; \theta_{h})$, where $\theta_{g}$ and $\theta_{h}$ are the disjoint parameter sets on which $g$ and $h$ are dependent, respectively.  Unless otherwise stated, the norm $||\ \cdot\ ||$ will refer to the $L_2$-norm. 





\begin{figure}
	\centering
	\includegraphics[width=0.45\linewidth]{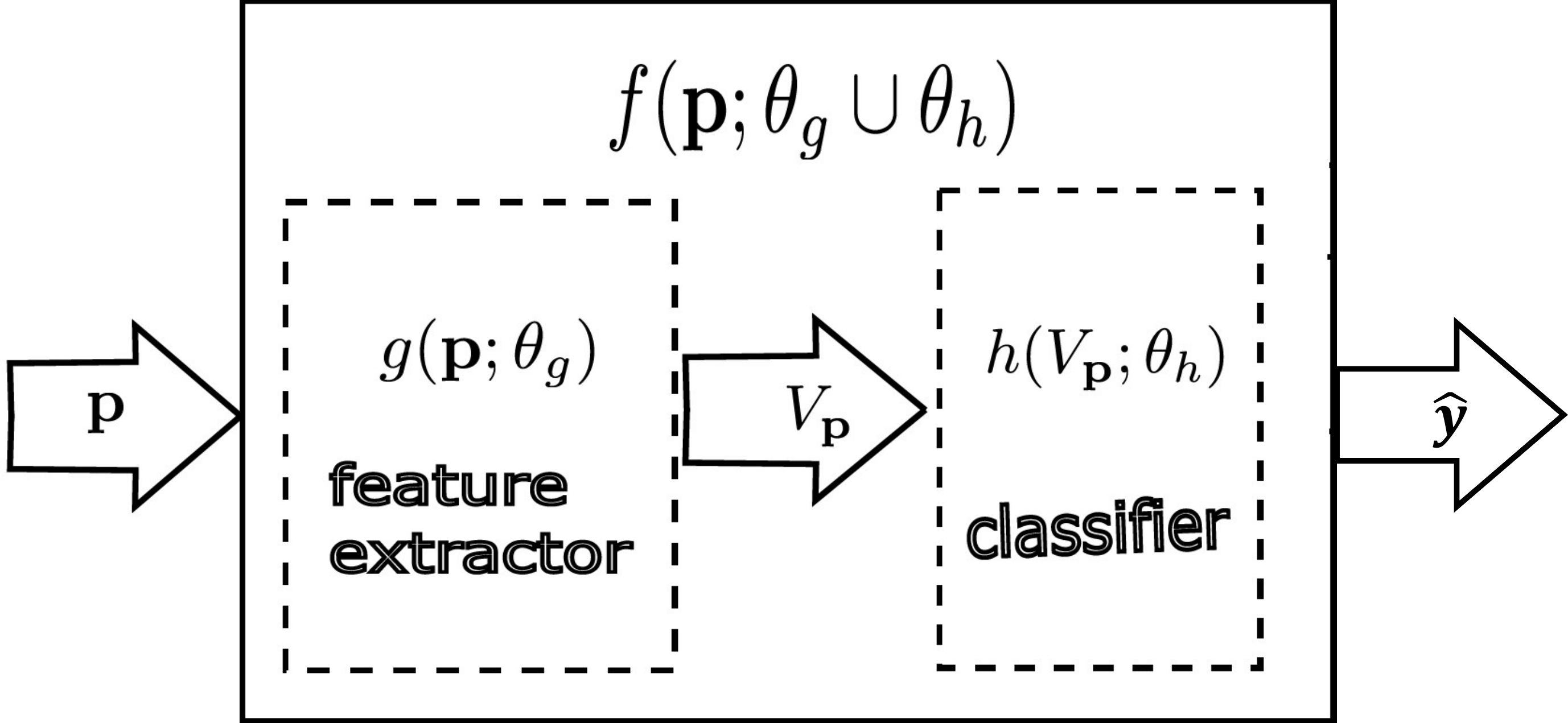}
	\caption[gm]{Generic Network showing it as a composition of feature extractor and a feature classifier.}
	\label{fig:NetworkGeneric}
\end{figure}

\paragraph{Prototype vector or image}
We envision the given neural network as split into three representations of the input as shown in Fig. \ref{fig:NetworkGeneric} : the input space, the feature space, and the output space. The fundamental data structure we employ to distinguish one class from another, not knowing its training examples, is the \emph{class prototype}.  We refer to a specific input vector $\vec{p}$ as a \emph{prototype} input.  Typically, there exists a separate prototype image for each class such as to create a prototype image set $\left\{\vec{p}^1, ... ,\vec{p}^K\right\}$.  A class prototype feature vector is the mapping of a prototype image to the feature layer of the network $V_p = g(\vec{p}; \theta_{g})$.  These features represent the important latent representations of a particular class' training set examples that the DNN trained on.

We wish to understand the performance of a trained neural network that is given to us without any data.  Although we do not have access to the original data, we find that in a classification setting the model parameters and outputs are sufficient to reveal important inter-class relationships whose quantities depend on 
\begin{inparaenum}[$a)$]
	\item the amount of training data used to train the model and 
	\item the generalization performance of the model. 
\end{inparaenum}
By exposing these inter-class relationships from the available information, we can then quantify them and make general statements about how successful the learning algorithm was.


To derive separate prototypes for each class, we try to enforce minimal overlap between different class' prototypes in the output space of the neural network; the one-hot encoded vectors for each class are the best mechanism to achieve this.  By establishing class prototypes that the network is maximally confident in, we can highlight the entire gambit of influential feature activations for each class.  Furthermore, if all the important features to a class are captured by the prototype, we can infer that there was at least one class example in the training set that utilized each prototypical feature; in this way our prototypes indirectly represent the training set even though we never see actual examples.  We now delve into the algorithms to create class prototype examples and the prototype-based metrics from which can infer network generalization performance.

\subsection{How to Create a Class Prototype Example}
\label{CreatePrototype}
We are given no data, but we do receive a trained, differentiable classifier $f(\vec{x};\theta)$ with full access to its weights.  In our approach, $\theta$ is fixed and we may sometimes omit it for succinctness.

To create a prototype example $\vec{p}$, we iteratively update the values in a randomly initialized input in $\R^N$ to minimize the cross-entropy loss between the output $f(\vec{p};\theta)$ and a target probability distribution $\vec{y}_p \in \R^K$ of our choosing.  Let $z$ refer to the iteration number and $\alpha$ to a chosen learning rate.  Then the update rule becomes:

To create a prototype example $\vec{p}$, we iteratively update the values in a randomly initialized input in $\R^N$ using Eqn.~\eqref{updateRule} to minimize the cross-entropy loss defined by Eqn.~\eqref{lossFunction} between the output $f(\vec{p};\theta)$ and a target probability distribution $\vec{y}_p \in \R^K$ of our choosing, where $z$ refer to the iteration number, $\alpha$ to a chosen learning rate, and $\mathcal{L}$, for a chosen $\vec{y}_p \in \R^K$.
\begin{align}
	\textrm{\textbf{Update Rule: }} &\vec{p}_{z+1} \leftarrow \vec{p}_{z} - \alpha\nabla_{\vec{p}}\mathcal{L}\ / ||\ \nabla_{\vec{p}}\mathcal{L}\ || \label{updateRule} \\
	\textrm{\textbf{Loss Function: }}&\mathcal{L}  = - \sum_k^{K} y_{p,k}\ log(f_k(\vec{p})) \label{lossFunction}
\end{align}





Effectively, we repeatedly forward pass the prototype example into the network, compute the cross-entropy loss of the prototype's output against a fixed probability vector, backpropagate, and update the prototype before the next iteration.  For a desired prototype example $\vec{p}^k$ for class k, we set $\vec{y}_p$ to the one-hot encoding for that class, i.e. each element of $\vec{y}_p$ is,
\begin{equation}\label{eq:3}
	y_{p,j}= 
	\begin{cases}
		1,& \text{if } j=k\\
		0, & \text{otherwise}
	\end{cases}
\end{equation}
The prototype example is then created by randomly initializing vector $\tilde{\vec{p}} \in [0,1]^N$ and then iteratively updating it with the update rule in Eqn. \ref{updateRule}.

\begin{figure}
	\centering
	\includegraphics[width=0.7\linewidth]{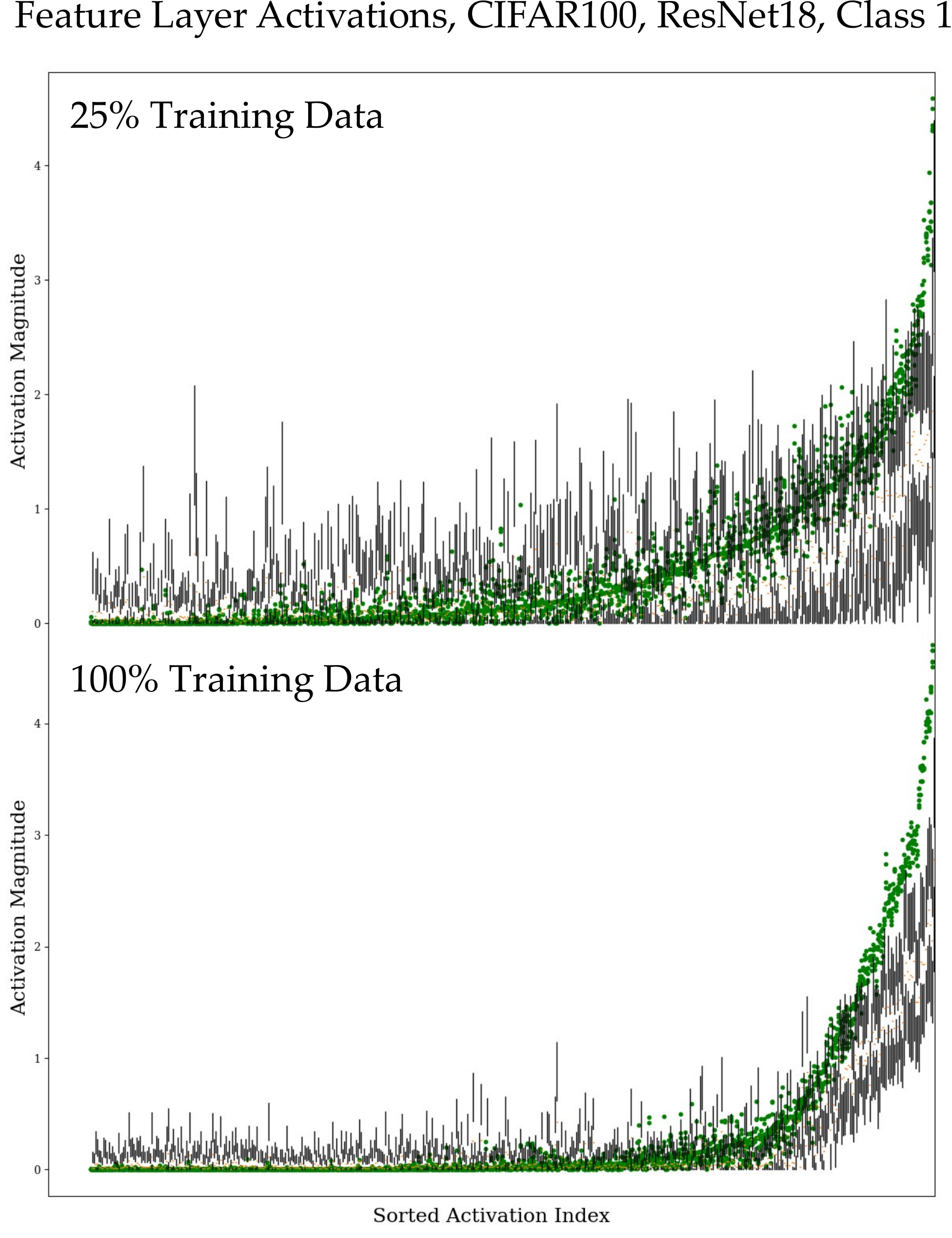}
	\caption[gm]{Plots of CIFAR100, ResNet18, class 1 activation levels (vertical axis) in the feature layer as a function of neuron index (horizontal axis).  Neuron indices are sorted by activation levels of the prototypes (green dots).  Prototype variation shows effect of 5 different random $\tilde{\vec{p}}$ initializations.  Black bars are interquartile ranges of class 1 training data.  (Top) Low quantity of training data. (Bottom) High quantity of training data.  Charts on same scale.}
	\label{fig:BoxWhisker}
\end{figure}

\subsection{Feature Layer Neuron Activation Profiles}
\label{NeuronActivationProfiles}
The box-and-whisker plots in Figure \ref{fig:BoxWhisker} show how a CIFAR100 class prototype comprehensively stimulates the most activated neurons for its class in the feature space (a 512x1 vector in this case).  In this figure, the activation levels (vertical axis) of all neurons in the feature layer are plotted against their respective vector indices, which are sorted by the activations of the prototype. The black vertical bars represent the interquartile ranges (IQR) of all the class' examples' feature activations from the training set.  The green dots represent 5 different versions of the class prototype that were generated from 5 different random initialization vectors.  The top and bottom charts differ in the amount of data the model had been trained on, with the top having been trained on only a small portion of data while the bottom model trained on all available data.  Both models were trained for 100 epochs and achieved 100\% training accuracy on their respective training sets.  

We highlight several important trends from these graphs:

\begin{itemize}
	\item In both graphs, the class prototype activations lie near or above the top of the IQRs from the class' training examples.  Particularly, for the most activated neurons near the right side of the horizontal axis, the prototype activations far exceed the training data, but only in the case of the well-trained 100\% data model do we see this excessive prototype activation appear over a wider range of neurons. 
	\item  The curves created by the sorted prototype data possess generally the same shape between charts, but the well-trained model exhibits a curve   
	\begin{inparaenum}[$a)$]
		\item that is tighter and less noisy between different random initializations of the prototype, 
		\item that has a sharper slope where the prototype activations transition from low activity neurons to high activity neurons in a almost piecewise-linear fashion, 
		and 
		\item that has more successfully suppressed a larger number of class-unimportant neurons;
	\end{inparaenum} the \emph{better} model relies on a fewer number of highly activated neurons and is able to dampen the response of the prototype and training examples in the non-activated region. 
	\item We have observed these behaviors in other classes.
	\item The \emph{better} model in the bottom graph had 17\% higher test accuracy.

\end{itemize}

Since \emph{a network's class prototype can be derived without data, and activation of neurons at the feature layer differ with the quality of the network's training (as observed in Fig.~\ref{fig:BoxWhisker}), we hypothesize that dataless metrics can be computed.} In this work we propose and validate two such metrics.

\begin{enumerate}[1.]
	\item \textbf{Proposed first metric, $\mathcal{M}_{g}$}, postulates that class prototypes should be less similar and utilize fewer of the same neurons in a higher performance model; harnessing the observations in bullet 2. 
	
	\item \textbf{Proposed second metric, $\mathcal{M}_{adv}$}, proposes that the nearest adversarial example
	(misclassification) of each class prototype will be further away in \emph{feature} space for a better trained model.  This idea stems from bullet 1, which observes the increased activation margin the class prototypes exhibit above their training data over a larger number of important class neurons.
\end{enumerate}

The fundamental measure of similarity we use for both metrics is the cosine similarity.  Given two vectors $\vec{v}_1$ and $\vec{v}_2$, the cosine similarity between the two is:

\begin{equation} \label{eq:4}
	cos(\theta) = \frac{\vec{v}_1 \cdot\ \vec{v}_2}{||\ \vec{v}_1\ ||\ ||\ \vec{v}_2\ ||}
\end{equation}


We select the magnitude independent cosine similarity measure as opposed to a raw $||\ \cdot\ ||_p$ norm distance since the overall prototype magnitudes and shapes could vary across classes.


\subsection {Computing $\mathcal{M}_{g}$}
\label{metric1}

Let the $k^{th}$ row of a matrix, $G \in \R^{K\text{x}K}$, be the unit feature vector of the class $k$ prototype: 


\begin{equation} \label{eq:5}
	G_k = g(\vec{p}^k)\ /\ ||\ g(\vec{p}^k)\ ||_2
\end{equation}

The elements of the matrix $GG^T \in [0,1]^{K\text{x}K}$ are the cosine similarities, see Eqn. \ref{eq:4}\ , of each pair of class prototype unit vectors:

\begin{equation} \label{eq:6}
	(GG^T)_{a,b} = \frac{g(\vec{p}^a)\cdot\ g(\vec{p}^b)}{||\ g(\vec{p}^a)\ ||\ ||\ g(\vec{p}^b)\ ||}
\end{equation}

The metric $\mathcal{M}_{g} \in [0,1]$ is then,

\begin{equation} \label{eq:7}
	\mathcal{M}_{g} = 1 - \frac{1}{K^2}\sum_{a,b}(GG^T)_{a,b}
\end{equation}

where we have computed the mean over all elements of $GG^T$ and subtracted this scalar from 1 to measure the average \emph{dissimilarity} between any two pairs of class prototypes in feature space.

\subsection{Computing $\mathcal{M}_{adv}$}\label{metric2}
\label{Metric2}
We desire to create a metric than can be useful in tracking generalization on clean examples, but also robust generalization.  Ideally, $\mathcal{M}_{adv}$ would scale proportionately with test accuracy on clean examples, but also inform us of differences in training algorithms between models we are comparing if one was standard cross-entropy trained and the other adversarially trained \cite{MadryPGD} to increase robustness.  Previous work on robust neural networks \cite{schmidt2018adversarially, TsiprasRobustnessOddsAccuracy} indicates that increasing data quantity aides in reducing the robust generalization gap, but for a fixed data set size, increasing robustness will reduce accuracy on clean examples.  We believe that we can capture both these effects by performing adversarial attacks \cite{chakraborty2018adversarial} on our class prototype images, which we postulate inherit the robustness characteristics of the network and their respective classes.  

A complication is that we desire to measure our metrics based on similarities in feature space on the interval [0,1], rather than based on unbounded $L_2$ distances in image space, which adversarial attacks are traditionally evaluated by.  We select the DeepFool \cite{DeepFool} attack as an efficient method for finding the closest adversary for each class prototype image and assume that close adversaries in the image space translate to acceptable solutions in the feature space.

Our computation of $\mathcal{M}_{adv}$ requires us to find the closest \emph{adversarial} example $\vec{p}_{adv} \in [0,1]^N$ for each class prototype and then compute each adversary's activation levels $\vec{V}_{p,adv}=g(\vec{p}_{adv})$.  By definition, an adversary is a perturbation field, $\delta$, applied to $\vec{p}$ that results in misclassification $y^k \ne \argmax_k(f(\vec{p}^k+\delta;\theta_f))$.  



Assume we perform the DeepFool algorithm on each class prototype, $\vec{p}^k$, to create a list of prototype input adversaries $\vec{p}_{adv}^k$ and forward pass them to find their feature vectors $\vec{V}_{p,adv}^k = g(\vec{p}_{adv}^k ; \theta_{g})$.  The computation of $\mathcal{M}_{adv}$ is then,

\begin{equation} \label{eq:9}
	\mathcal{M}_{adv} = 1 - \frac{1}{K}\sum_{k}^{K}\frac{g(\vec{p}^k)\cdot\ g(\vec{p}_{adv}^k)}{||\ g(\vec{p}^k)\ ||\ ||\ g(\vec{p}_{adv}^k)\ ||}
\end{equation}

where we calculate the average cosine similarity, Eqn. \ref{eq:4}, between all class prototypes and their respective DeepFool adversaries in feature space.  This value is subtracted from 1 to convey the average dissimilarity.

\section{Empirical Evaluation of Proposed Metrics}
\label{EmpiricalEvaluationOfDNNs}

Our experimental goal was to see if proportionality trends existed between the generalization test accuracy of a deep neural network on common image classification tasks and our dataless metrics.  After an exhaustive search and many computations, $\mathcal{M}_g$ and $\mathcal{M}_{adv}$ are two of the best metrics we found that successfully inferred test accuracy.  We are presenting their results for one standard cross-entropy trained architecture and three datasets, but we plan on adapting them to other architectures, datasets, and tasks in the future.  We instantiate our generic network in Fig. \ref{fig:NetworkGeneric} as a ResNet18 \cite{ResNet}, defining $g(\ \cdot\ ;\theta_{g})$ as the input to the flattened layer after global average pooling and $h(\ \cdot\ ;\theta_{h})$ as the fully connected and softmax layers.  All evaluations were completed on a single GPU.


 
\paragraph{Datasets}

We focused on the image classification datasets CIFAR10 \cite{CIFAR10}, CIFAR100 \cite{CIFAR10}, and Tiny ImageNet \cite{tinyImage}.  All of these datasets contain labeled 3-channel color images of natural objects such as animals and transportation vehicles.  The datasets differ in their image sizes (32x32 pixels for CIFAR and 64x64 for Tiny), in number of categories (10, 100, and 200 for CIFAR10, CIFAR100, and Tiny ImageNet respectively), and the number of training examples per category (5000 per class in CIFAR10 and 500 per class for CIFAR100/Tiny ImageNet).

\paragraph{Training}

\begin{figure}
	\centering
	\includegraphics[width=0.9\linewidth]{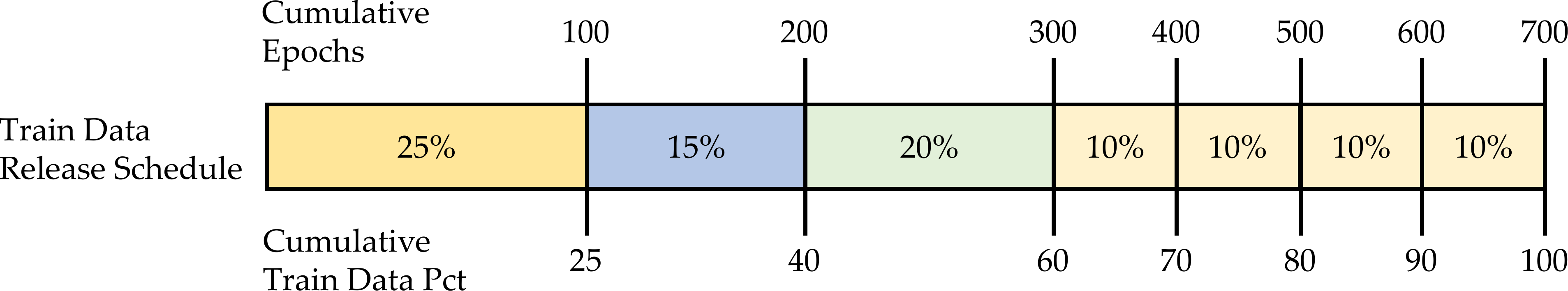}
	\caption[gm]{Percentage of training data released to deep neural network as a function of cumulative epochs of training.  Metrics and test accuracy computed at each 100 epochal increment checkpoint.}
	\label{fig:DataSchedule}
\end{figure}

For each dataset, we randomly initialize a ResNet18 \cite{ResNet} network and train it for a total of 700 epochs.  However, as shown in Fig. \ref{fig:DataSchedule}, we gradually increment the fraction of the full training set available to the network every 100 epochs, in a class-stratified fashion, beginning with only 25\% of the training data in the first 100 epochs.  At every 100 epoch checkpoint, we measure the test accuracy of the network, generate protoypes for each class given the current state of the model, and compute our metrics.  The graphs and tables in this section reflect these measurements.  The learning rate starts at 0.1 in the first 100 epochs, but begins at 0.05 for epochs 100-700, following a cosine annealing schedule for each 100 epochs.  With the network frozen, prototype images are computed per Eqn. \ref{updateRule} with a learning rate of 0.1 (CIFAR100,Tiny) or 0.01 (CIFAR10).
\paragraph{Overall}
Table \ref{TableResults} shows the detailed epochal checkpoint results for each training data percentage, dataset, model accuracy, and metric.  In general, the metrics are highly correlated and trend proportionately with model test accuracy.  The accuracy results reflect a single model initialization undergoing the 700 epoch recipe in Fig. \ref{fig:DataSchedule}, but the metric results are the mean of 5 different prototype sets.  Since the metrics themselves are the mean of $K^2$ ($\mathcal{M}_g$) and $K$ ($\mathcal{M}_{adv}$) elements, a single number in the table corresponds to a mean of means.

\begin{table*} 
\begin{center}
	\caption{Test Accuracy and Metric Results}
	\label{TableResults}
	\begin{tabular}{lccccccc}\toprule
		Training Data Percent & 0.25 & 0.4 & 0.6 & 0.7 & 0.8 & 0.9 & 1.0 \\\midrule
		CIFAR100 & & &  & & & &\\ [.1cm]
		Test Accuracy & .556 & .596 & .647 & .664 & .684 & .706 & .723 \\ [.1cm]
		Metric 1 & .707 & .717 & .738 & .762  & .786  & .796  & .811  \\ [.1cm]
		Metric 2 & .306 & .317 & .342 & .372  & .396  & .401  & .426 \\\midrule
		Tiny ImageNet &  &  &  &  &   &   &  \\[.1cm]
		Test Accuracy & .408 & .442 & .476 & .490 & .509 & .524 & .531 \\[.1cm]
		Metric 1 & .709 & .727 & .750 & .790 & .816 & .829 & .845 \\[.1cm]
		Metric 2 & .307 & .308 & .327 & .376 & .429 & .459 & .495 \\\midrule
		CIFAR10 &  &  &  &  &   &   &  \\[.1cm]
		Test Accuracy & .866 & .885 & .906 & .917 & .925 & .934 & .940 \\[.1cm]
		Metric 1 & .783 & .794 & .812 & .822 & .823 & .826 & .829 \\[.1cm]
		Metric 2 & .323 & .293 & .334 & .351 & .391 & .371 & .372 \\\bottomrule
	\end{tabular}
	\stoptable{Results averaged over 5 sets of random prototype initializations.}
\end{center}
\end{table*}

\subsection{Evaluation of Metric $\mathcal{M}_{g}$}
\label{EvalMetic1}
\begin{figure}[htb]

	\begin{subfigure}[b]{0.32\textwidth}
		\centering
		\includegraphics[width=\textwidth]{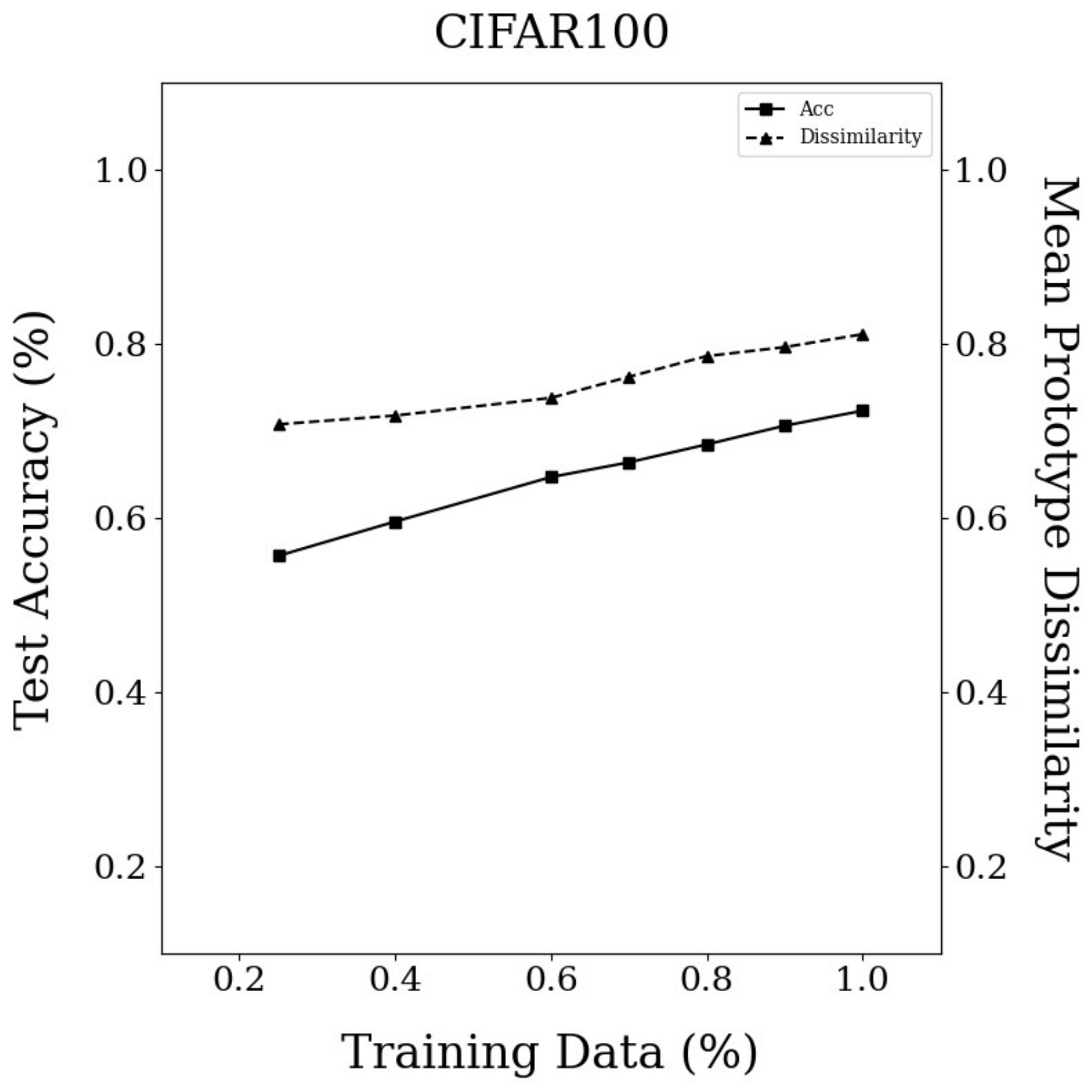}
		\caption{CIFAR100}
		\label{c100a}
	\end{subfigure}
	\hfill
	\begin{subfigure}[b]{0.32\textwidth}
		\centering
		\includegraphics[width=\textwidth]{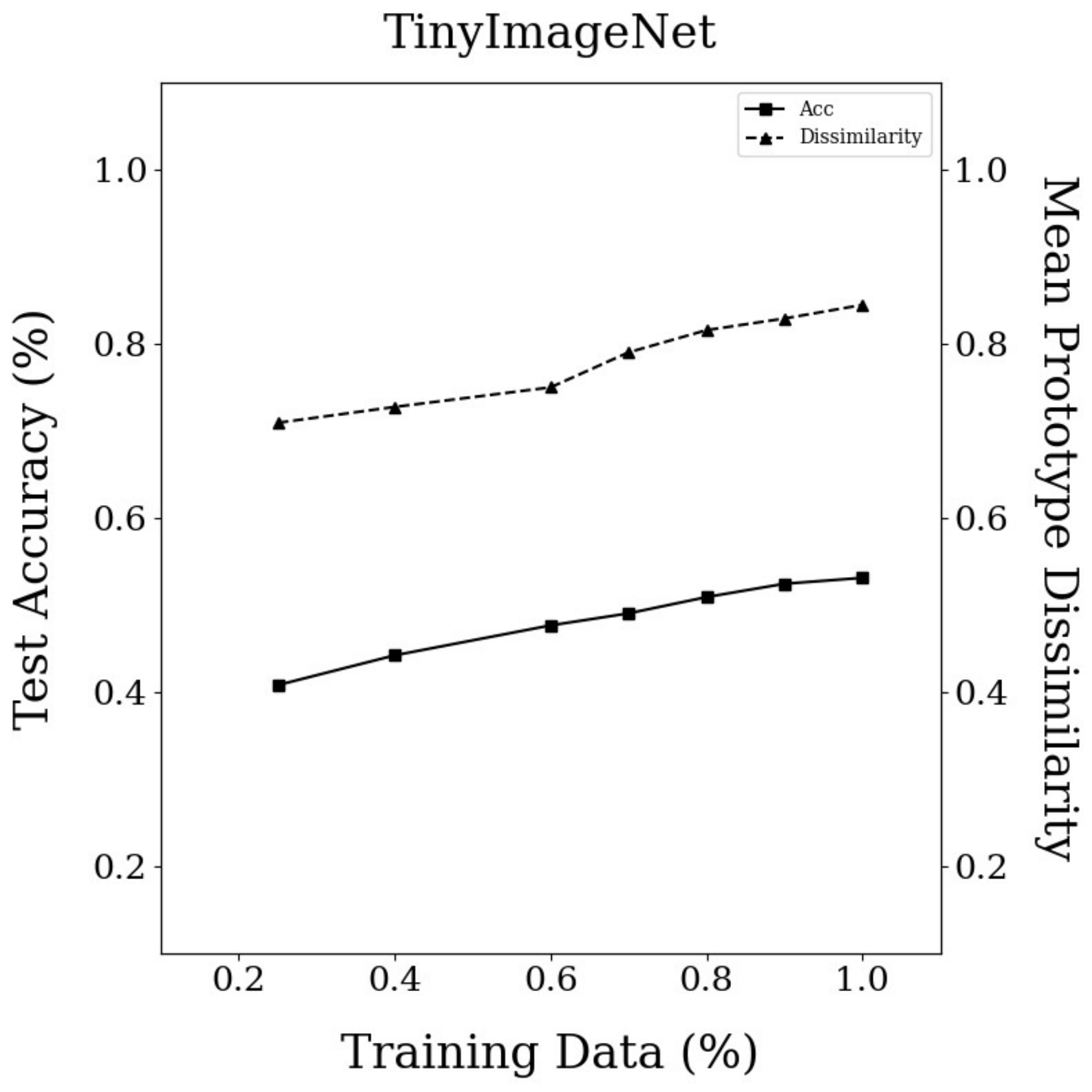}
		\caption{Tiny ImageNet}
		\label{tinya}
	\end{subfigure}
	\hfill
	\begin{subfigure}[b]{0.32\textwidth}
		\centering
		\includegraphics[width=\textwidth]{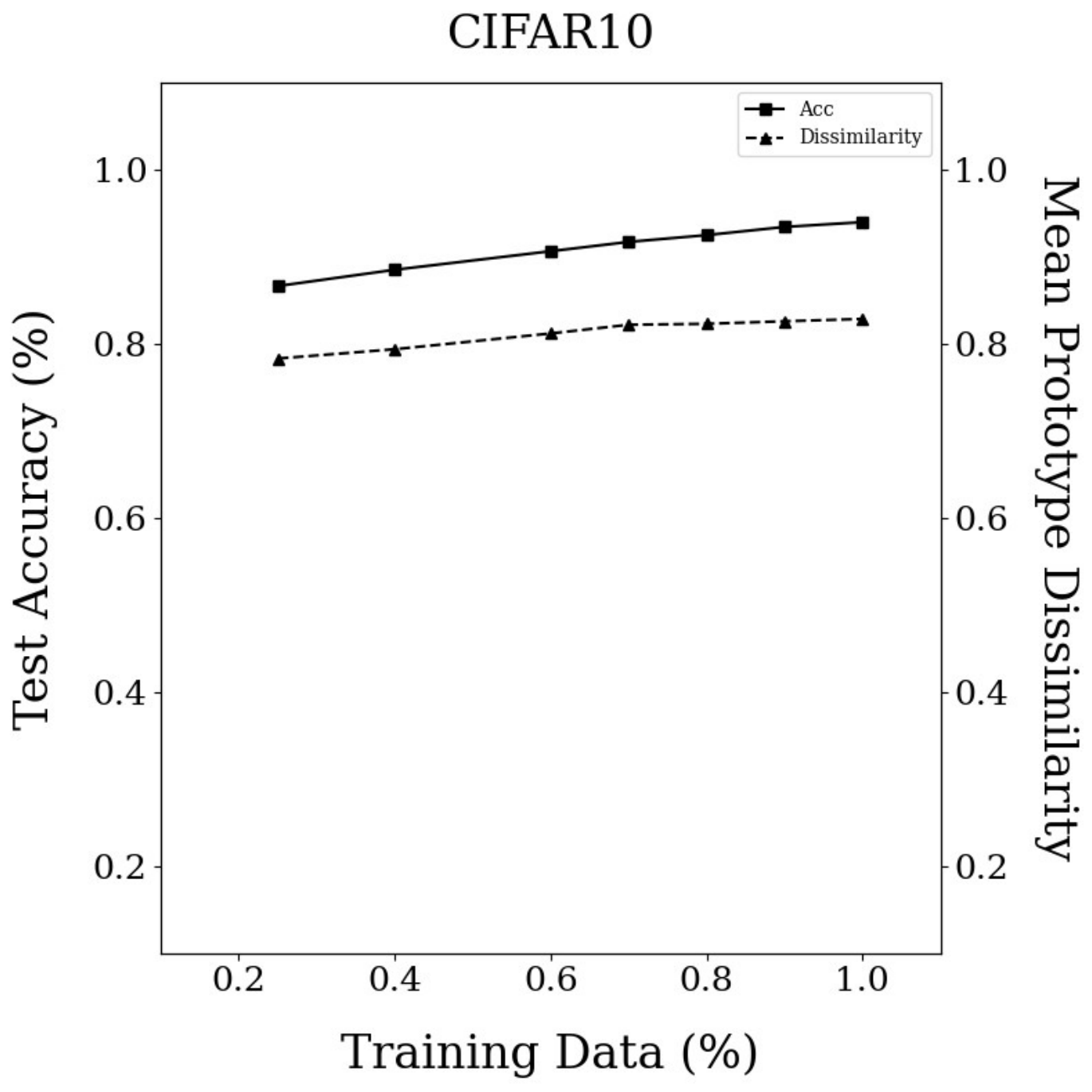}
		\caption{CIFAR10}
		\label{c10a}
	\end{subfigure}
	\hfill
	\caption[gm]{$\mathcal{M}_{g}$ - Plots of test accuracy and mean prototype dissimilarity ($\mathcal{M}_{g}$) as a function of training data fraction used in training.}
	\label{fig:Mga}
\end{figure}

\begin{figure}[htb]
	
	\begin{subfigure}[b]{0.3\textwidth}
		\centering
		\includegraphics[width=\textwidth]{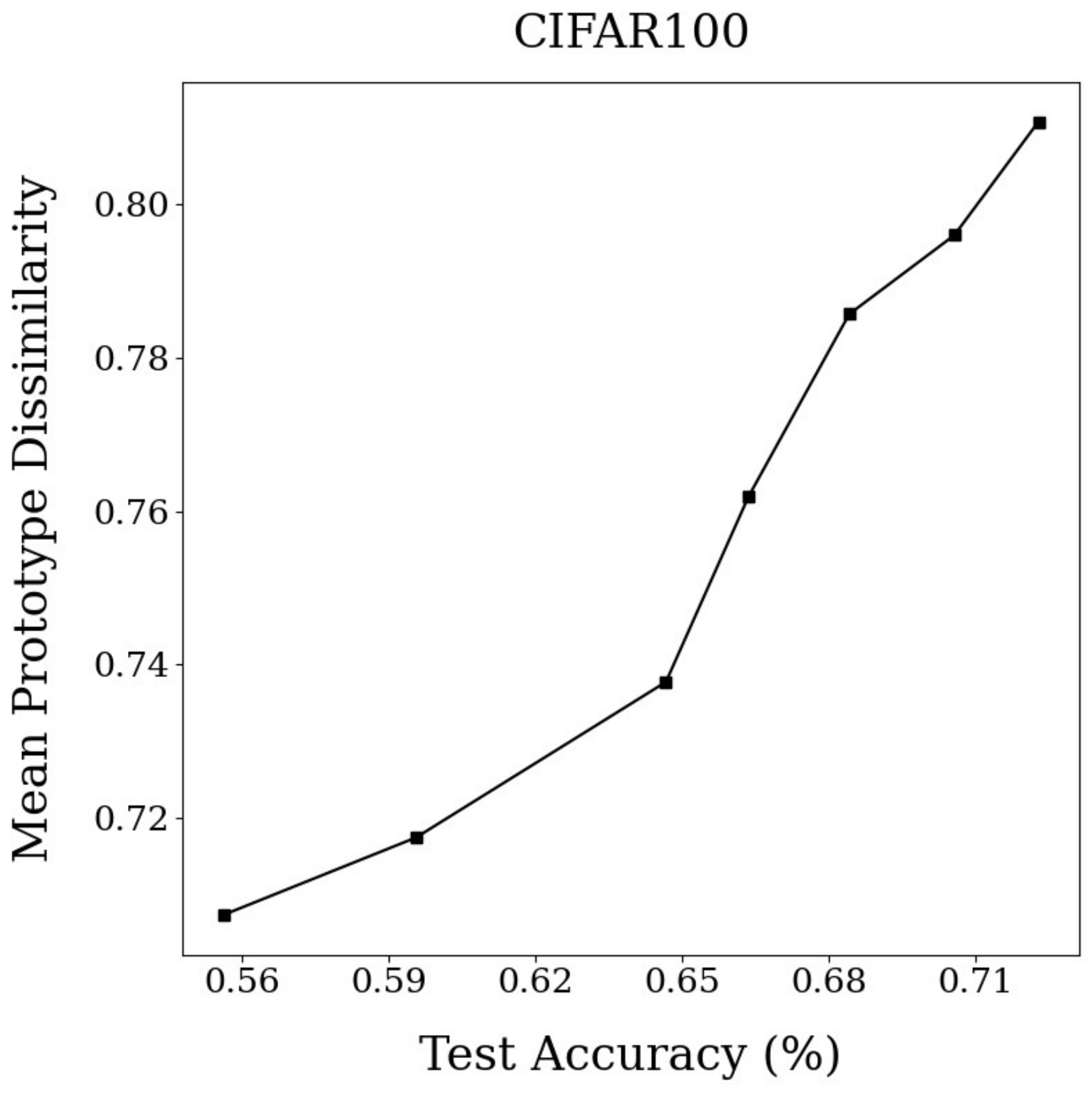}
		\caption{CIFAR100}
		\label{c100b}
	\end{subfigure}
	\hfill
	\begin{subfigure}[b]{0.3\textwidth}
		\centering
		\includegraphics[width=\textwidth]{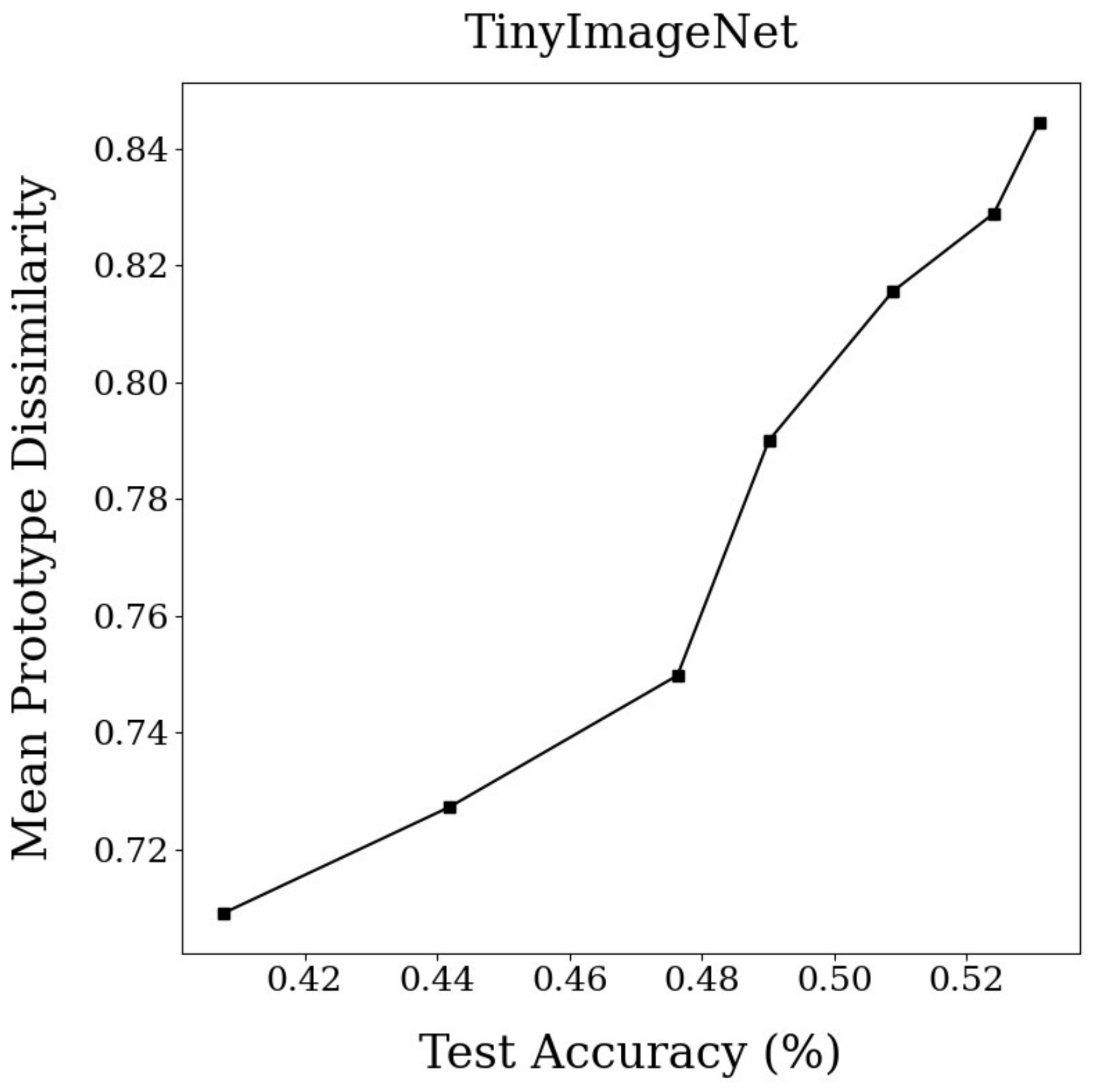}
		\caption{Tiny ImageNet}
		\label{tinyb}
	\end{subfigure}
	\hfill
	\begin{subfigure}[b]{0.3\textwidth}
		\centering
		\includegraphics[width=\textwidth]{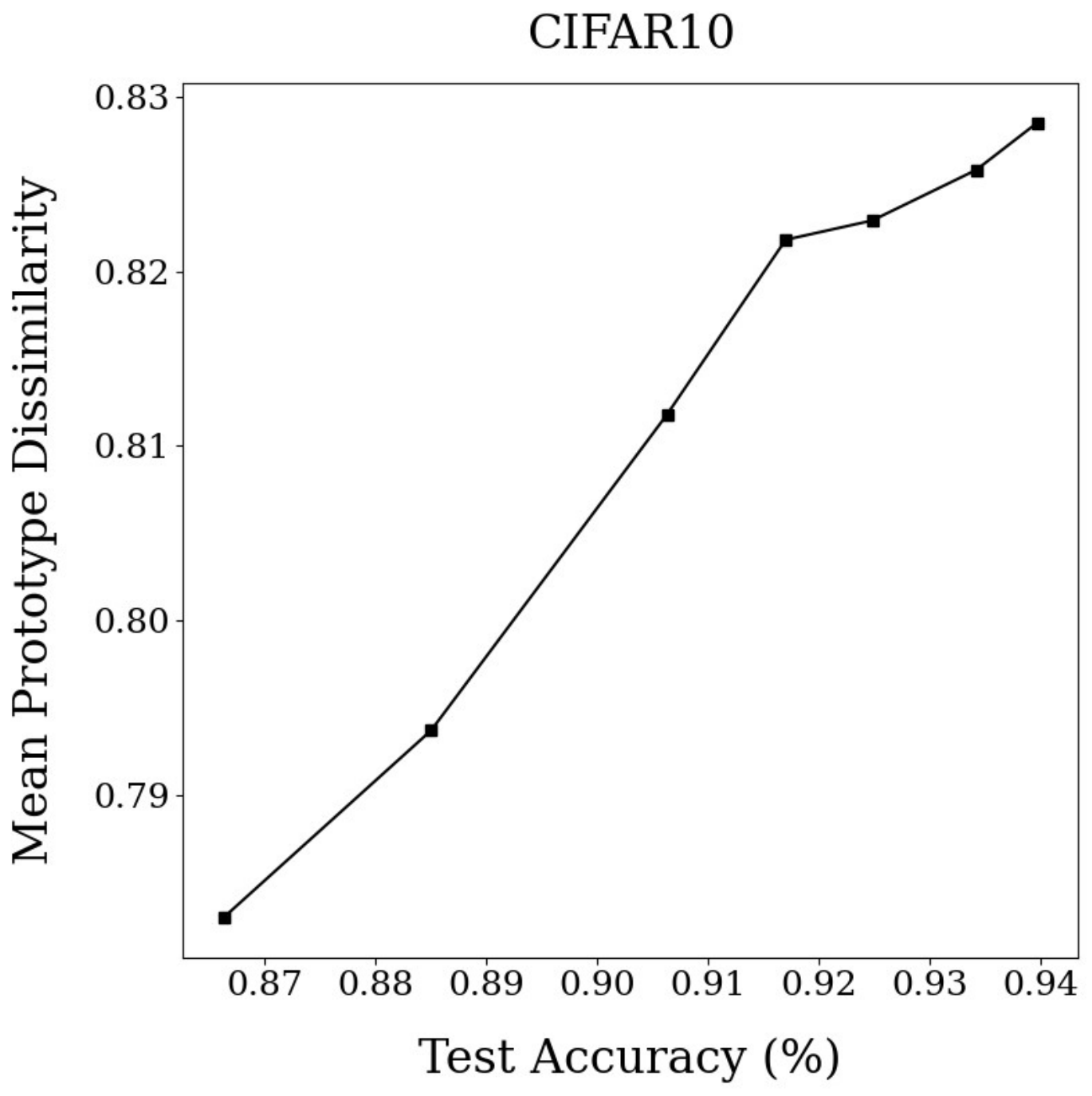}
		\caption{CIFAR10}
		\label{c10b}
	\end{subfigure}
	\hfill
	\caption[gm]{$\mathcal{M}_{g}$ - Plots of mean prototype dissimilarity ($\mathcal{M}_{g}$) as a function of test accuracy.  Pearson correlation coefficients 0.97 (CIFAR100), 0.97 (Tiny), and 0.99 (CIFAR10).}
	\label{fig:Mgb}
\end{figure}

Figure \ref{fig:Mga} shows trendlines for test accuracy and metric $\mathcal{M}_g$ as a function of available training data percentage for the three datasets.  As the test accuracy increases, $\mathcal{M}_g$ increases in a proportional manner throughout the training data percentage range.  Figure \ref{fig:Mgb} eliminates the training data percentage variable and plots our two variables of interest, $\mathcal{M}_g$ and test accuracy, directly against each other.  From these charts, we created best-fit linear lines and measured high Pearson correlation coefficients of 0.97, 0.97, and 0.99 between $\mathcal{M}_g$ and test accuracy for CIFAR100, Tiny ImageNet, and CIFAR10, respectively.




\subsection{Evaluation of Metric $\mathcal{M}_{adv}$}
\label{EvalMetic2}
\begin{figure}[htb]
	
	\begin{subfigure}[b]{0.3\textwidth}
		\centering
		\includegraphics[width=\textwidth]{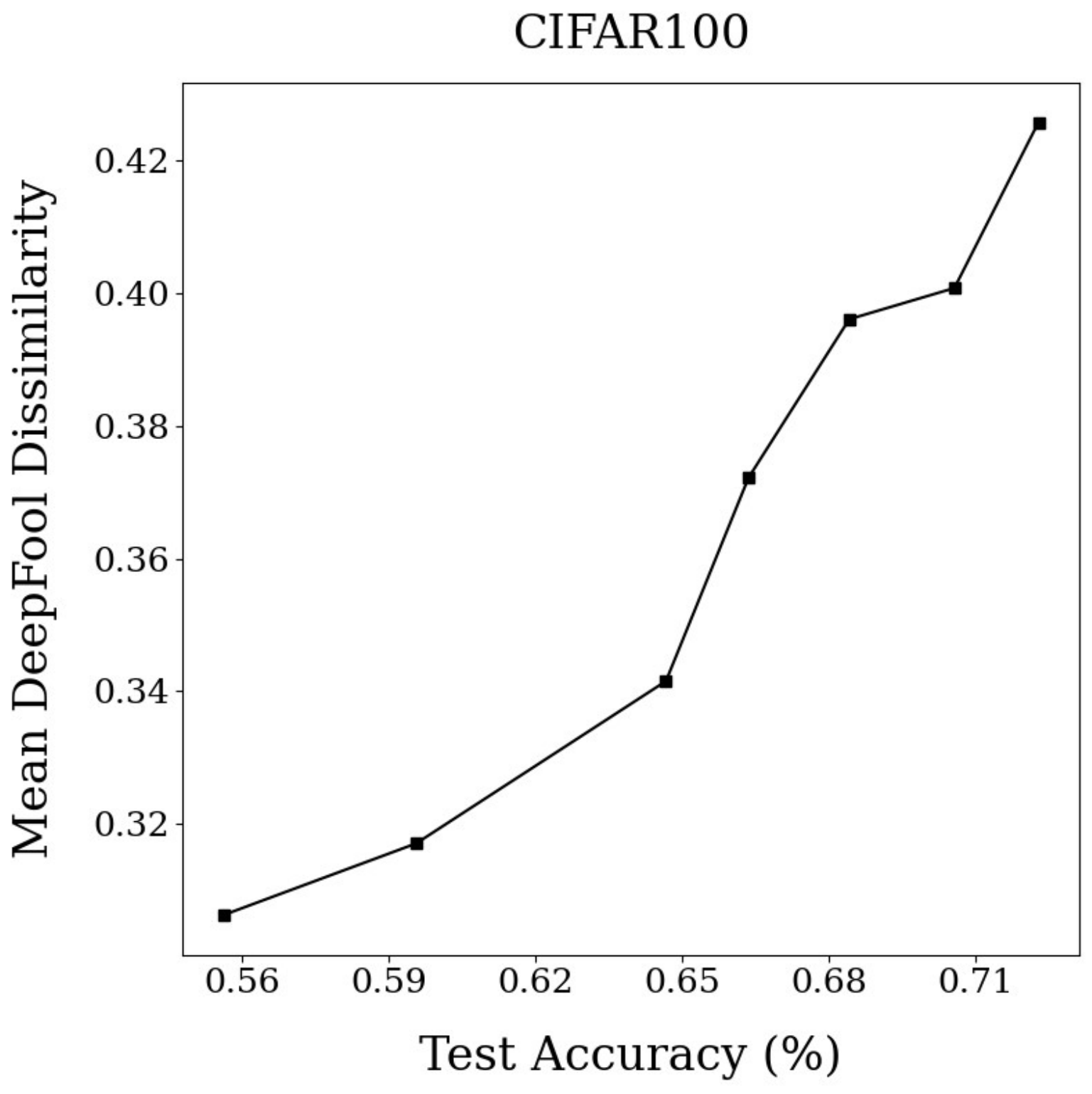}
		\caption{CIFAR100}
		\label{c100advb}
	\end{subfigure}
	\hfill
	\begin{subfigure}[b]{0.3\textwidth}
		\centering
		\includegraphics[width=\textwidth]{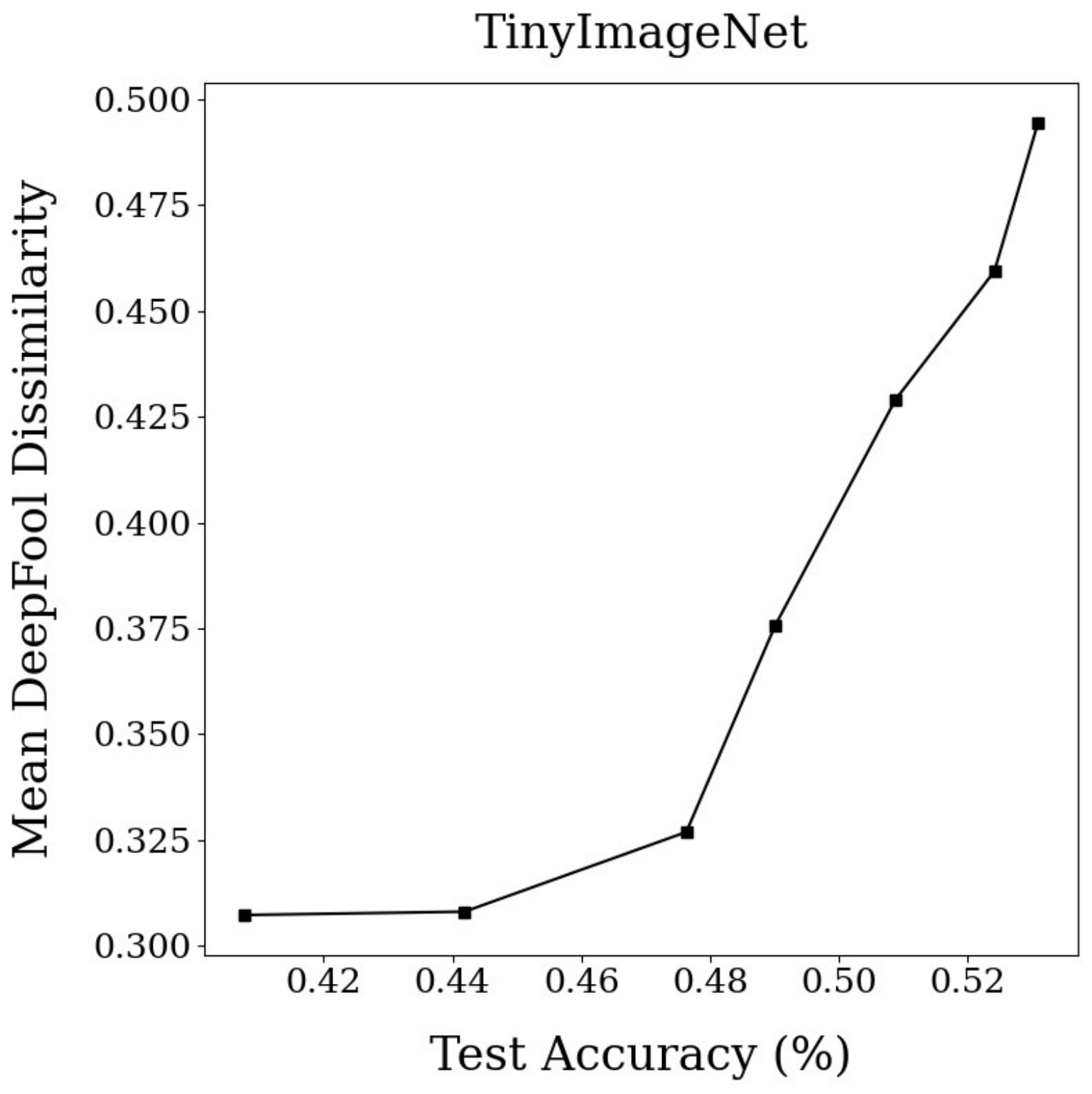}
		\caption{Tiny ImageNet}
		\label{tinyadvb}
	\end{subfigure}
	\hfill
	\begin{subfigure}[b]{0.3\textwidth}
		\centering
		\includegraphics[width=\textwidth]{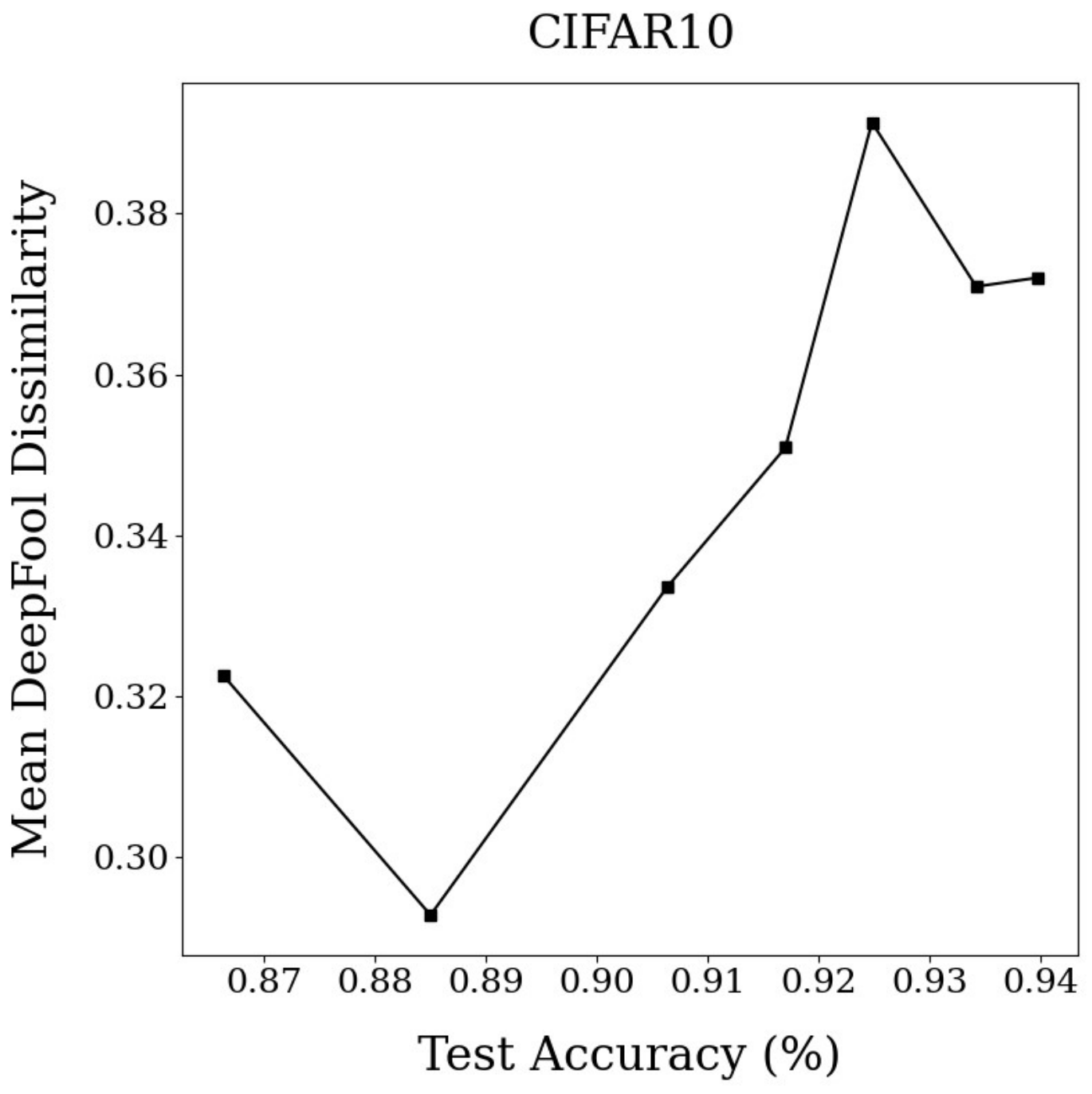}
		\caption{CIFAR10}
		\label{c10advb}
	\end{subfigure}
	\hfill
	\caption[gm]{$\mathcal{M}_{adv}$ - Plots of mean prototype DeepFool adversary dissimilarity ($\mathcal{M}_{adv}$) as a function of test accuracy.}
	\label{fig:Madvb}
\end{figure}

Figure \ref{fig:Madvb} plots metric $\mathcal{M}_{adv}$ as a function of test accuracy for all three datasets.  In this setting, we again computed the Pearson correlation coefficients, which although less correlated than for $\mathcal{M}_{g}$, are still quite high with 0.97, 0.91, and 0.82 for CIFAR100, Tiny ImageNet, and CIFAR10, respectively.


The behavior of $\mathcal{M}_{adv}$ in the low data regime is less well behaved, being noisy for CIFAR10 and effectively constant for Tiny ImageNet.  We attribute the low-data behavior of this metric in all three datasets to the model relying on a larger group of noisy features for each class (see Fig. \ref{fig:BoxWhisker}).  Only when the model is able to define a tighter group of less noisy features for each class (the bottom chart of Fig. \ref{fig:BoxWhisker}), do we see the dissimilarity between prototypes and their adversaries begin increasing.

In future work, we plan to evaluate this metric further to see if it is able to capture differences in generalization between non-robust and robust networks.

\section{Application of Proposed Methods}
\label{Applications}
We envision two immediate applications of the proposed method for dataless evaluation of DNNs. It can be used to compare performances of two networks or to evaluate performance of any DNN without using training, validation, or testing data.

\paragraph{Evaluating a DNN}
\label{sec:ComparisonOfTwoNetworks}
Section~\ref{DatalessPerformanceEvalation} has developed and described steps for evaluating a DNN. Section~\ref{EmpiricalEvaluationOfDNNs} has used the proposed method to evaluate three trained DNNs. In this paragraph we outline how one should utilize the proposed method. Let us assume that a DNN is trained to classify an input to one of $K$ categories, which are given truth labels as one-hot encodings.  
\begin{enumerate}
	\item[Step 1:] Using the method described in Section~\ref{CreatePrototype} create $K$ prototype inputs, one for each category.
	\item[Step 2:] Using the method described in Section~\ref{metric1} compute the DNN's $\mathcal{M}_g$ value. Higher the value of $\mathcal{M}_g$, the better the expected performance with the real data. Note that, $0 \leq \mathcal{M}_g \leq 1$. Our evaluation of this metric indicates that a well trained DNN should have a value of 0.8 or higher.
	\item[Step 3:] Using the method described in Section~\ref{metric2} compute the DNN's  $\mathcal{M}_{adv}$ value. Again, higher the value of $\mathcal{M}_{adv}$, the higher the robustness against adversarial attacks. 
	Note that, $0 \leq \mathcal{M}_{adv} \leq 1$. Our evaluation of this metric indicates that a well trained DNN should have a value of 0.35 or higher.
\end{enumerate}

\paragraph{Comparing  DNNs}
\label{sec:ComparisonOfTwoNetworks}
Suppose that two or more DNNs are designed, trained, validated, and tested for the same task on custom curated datasets  that cannot be made available to the users of the DNNs because of proprietary  or privacy nature of the data.  All these DNNs' weights are available to an end-user. For selecting the best network the user can utilize the proposed dataless DNN evaluation method proposed here. For each of the DNN, the user should calculate the two metrics (following the steps described above) and compare the values to select the one that serves the user the best. If the user's priority is the higher accuracy, the user should select the DNN that has highest $\mathcal{M}_g$. On the other hand if the user is concerned more with adversarial attacks than the higher accuracy, the user should choose the DNN that has the highest $\mathcal{M}_{adv}$ value.

\section{Conclusion}
\label{Conclusion}
In this work we have proposed a method for datalessly evaluating the performance of DNNs. The proposed method works for evaluation of DNNs performing classification tasks. Given a trained DNN, the proposed method can create a `prototype' example for each category. Then  these prototypes are used to calculate two metrics: one for measuring classification accuracy and the other for robustness against adversarial attacks. While it is possible that a network may score highest in both measures, there are networks that would do better in one but not in the other.
\par
We have used the proposed methods to evaluate many DNN models. In this paper we report evaluation of three DNN models, developed using RestNet18 architecture to classify CIFAR10, CIFAR100, and Tiny ImageNet. The evaluations have proven quite effective; we are now trying to improve the method further.
\par
 Our long-term goal is for the idea of dataless metrics to extend beyond image classification to other tasks such as object detection and generative applications.  The core idea is that the model parameters alone can inform us of their ability to complete the task they were trained for.  We believe that dataless evaluations of intelligent models created using machine learning is critical to the long-term democratization of the field.
\pagebreak
\bibliographystyle{plain}
\bibliography{citations_proto,citations,DsBib}



\end{document}